\documentclass[10pt,twocolumn,letterpaper]{article}

\usepackage{cvpr}              

\usepackage[dvipsnames]{xcolor}

\definecolor{cvprblue}{rgb}{0.21,0.49,0.74}
\definecolor{mygray}{gray}{.9}
\usepackage[pagebackref,breaklinks,colorlinks,citecolor=cvprblue]{hyperref}
\usepackage[accsupp]{axessibility}

\def\paperID{10142} 
\def\confName{CVPR}
\def\confYear{2024}

\title{All in One Framework for Multimodal Re-identification in the Wild}

\author{He Li$^{1}$  \quad Mang Ye$^{1}$\thanks{Corresponding Author: Mang Ye (yemang@whu.edu.cn)}  \quad Ming Zhang$^{2}$  \quad Bo Du$^{1}$\\
$^1$National Engineering Research Center for Multimedia Software, Institute of Artificial Intelligence,\\
School of Computer Science, Hubei Luojia Laboratory,Wuhan University, Wuhan, China,\\
$^2$Guangzhou Urban Planning Design Survey Research Institute, Guangzhou, China
}

\begin{document}
\maketitle
\begin{abstract}
In Re-identification (ReID), recent advancements yield noteworthy progress in both unimodal and cross-modal retrieval tasks. However, the challenge persists in developing a unified framework that could effectively handle varying multimodal data, including RGB, infrared, sketches, and textual information.
Additionally, the emergence of large-scale models shows promising performance in various vision tasks but the foundation model in ReID is still blank. 
In response to these challenges, a novel multimodal learning paradigm for ReID is introduced, referred to as  All-in-One (AIO), which harnesses a frozen pre-trained big model as an encoder, enabling effective multimodal retrieval without additional fine-tuning.
The diverse multimodal data in AIO are seamlessly tokenized into a unified space, allowing the modality-shared frozen encoder to extract identity-consistent features comprehensively across all modalities.
Furthermore, a meticulously crafted ensemble of cross-modality heads is designed to guide the learning trajectory.
AIO is the \textbf{first} framework to perform all-in-one ReID, encompassing four commonly used modalities. Experiments on cross-modal and multimodal ReID reveal that AIO not only adeptly handles various modal data but also excels in challenging contexts, showcasing exceptional performance in zero-shot and domain generalization scenarios.
Code will be available at: \url{https://github.com/lihe404/AIO}.
\end{abstract}

\begin{figure}
    \centering
    \includegraphics[width=0.96\linewidth]{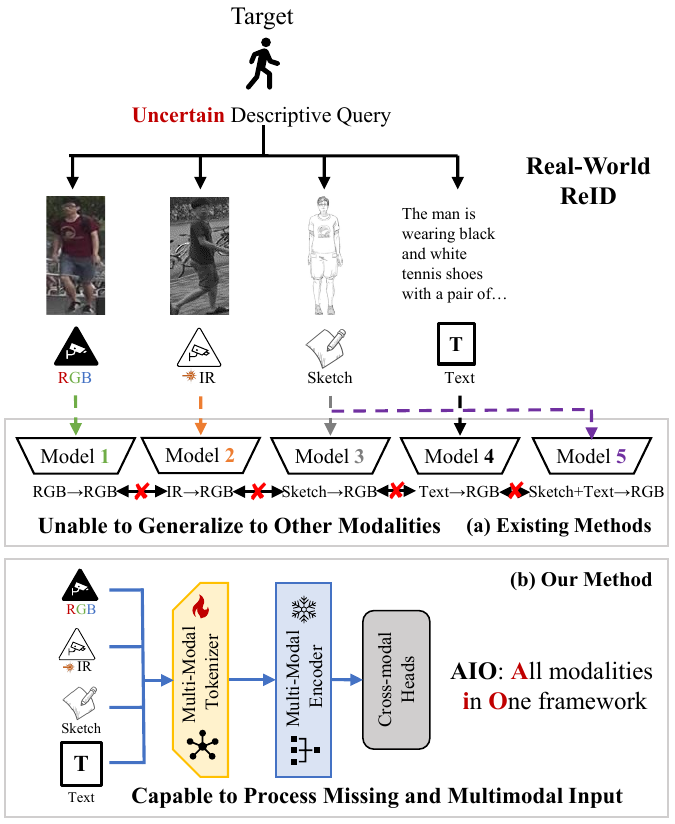}
    \vspace{-3mm}
    \caption{\textbf{Illustration of the proposed AIO and existing methods.} (a) Existing ReID methods \cite{ye2020dynamic,zhu2021dssl,chen2023modalityagnostic} independently learn the cross-modal ReID models, incapable of handling the uncertain input modalities in real-world scenarios.  (b) Our proposed AIO framework exhibits the capability to proficiently manage diverse combinations of input modalities, thus addressing the inherent uncertainties prevalent in practical deployment scenarios.}
    \label{fig:comparison}
    \vspace{-3mm}
\end{figure}

\section{Introduction}
\label{sec:intro}
Person Re-identification (ReID) aims to retrieve a target person captured by multiple non-overlapping cameras \cite{ye2021deep}. It is widely used in intelligent surveillance, security, and many other fields. ReID has been deeply studied in recent years and achieves human-level performance in both unimodal and cross-modal retrieval tasks \cite{ye2020dynamic,bao2023learning,zhang2022fmcnet,huang2023generalization}. 

Existing works are capable of retrieving between RGB images or leveraging different modalities of the query (\eg infrared (IR), sketch, or text) to find the person in RGB images \cite{jiang2023crossmodal,ye2024channel,fujii2023bilma,yang2023grand,huang2024federated}. 
However, RGB images are susceptible to environmental light fluctuations, while IR and sketch images lack vital color information crucial for ReID tasks. 
The adage ``\textit{a picture is worth a thousand words}" underscores the ease of accessing textual information, though it falls short in providing intricate visual details \cite{shao2022learning}. 
Moreover, as shown in \cref{fig:comparison}(a) and \cref{tab:comparison}, existing cross-modal methods are confined to specific paired modalities and models during training, rendering them unable to handle diverse input modalities effectively. Consequently, the generalizability of these methods to unseen modalities is severely hampered, a significant hurdle given the uncertainty of the modality in real-world query scenarios. This lack of adaptability across modalities severely constrains the practical applicability of existing methods in a practical real-world deployment with uncertain testing environments. Thus, \textit{1) How to improve the generalizability of modality is a significant challenge?}

\begin{table}[t]\small
    \centering
    \begin{tabular}{r||c|c|c|c|c}
        \thickhline
        \rowcolor{mygray} Method & RGB & IR & Sketch & Text & Multi\\
        \hline\hline
        NFormer \cite{wang2022nformer} & \color{LimeGreen}\Checkmark & \color{BrickRed}\XSolidBrush & \color{BrickRed}\XSolidBrush & \color{BrickRed}\XSolidBrush & \color{BrickRed}\XSolidBrush\\
        DC-Former \cite{li2023dcformer} & \color{LimeGreen}\Checkmark & \color{BrickRed}\XSolidBrush & \color{BrickRed}\XSolidBrush & \color{BrickRed}\XSolidBrush & \color{BrickRed}\XSolidBrush\\
        AGW \cite{ye2021deep} & \color{LimeGreen}\Checkmark & \color{LimeGreen}\Checkmark & \color{BrickRed}\XSolidBrush & \color{BrickRed}\XSolidBrush & \color{BrickRed}\XSolidBrush\\
        DART \cite{yang2022learning} & \color{LimeGreen}\Checkmark & \color{LimeGreen}\Checkmark & \color{BrickRed}\XSolidBrush & \color{BrickRed}\XSolidBrush & \color{BrickRed}\XSolidBrush\\
        Gui \etal \cite{gui2020learning} & \color{LimeGreen}\Checkmark & \color{BrickRed}\XSolidBrush &  \color{LimeGreen}\Checkmark & \color{BrickRed}\XSolidBrush & \color{BrickRed}\XSolidBrush\\
        SketchTrans \cite{chen2022sketch} & \color{LimeGreen}\Checkmark & \color{BrickRed}\XSolidBrush &  \color{LimeGreen}\Checkmark & \color{BrickRed}\XSolidBrush & \color{BrickRed}\XSolidBrush\\
        APTM \cite{yang2023unified} & \color{LimeGreen}\Checkmark & \color{BrickRed}\XSolidBrush & \color{BrickRed}\XSolidBrush & \color{LimeGreen}\Checkmark & \color{BrickRed}\XSolidBrush\\
        BiLMa \cite{fujii2023bilma} & \color{LimeGreen}\Checkmark & \color{BrickRed}\XSolidBrush & \color{BrickRed}\XSolidBrush & \color{LimeGreen}\Checkmark & \color{BrickRed}\XSolidBrush\\
        TriReID \cite{zhai2022trireid} & \color{LimeGreen}\Checkmark & \color{BrickRed}\XSolidBrush & 
        \color{LimeGreen}\Checkmark & \color{LimeGreen}\Checkmark & \color{LimeGreen}\Checkmark \color{black}(3)\\
        UNIReID \cite{chen2023modalityagnostic} & \color{LimeGreen}\Checkmark & \color{BrickRed}\XSolidBrush & 
        \color{LimeGreen}\Checkmark & \color{LimeGreen}\Checkmark & \color{LimeGreen}\Checkmark \color{black}(3)\\
        \hline\hline
        AIO & \color{LimeGreen}\Checkmark & \color{LimeGreen}\Checkmark & \color{LimeGreen}\Checkmark & \color{LimeGreen}\Checkmark & \color{LimeGreen}\Checkmark \color{black}(4)\\
        \thickhline
    \end{tabular}
    \vspace{-3mm}
    \caption{\textbf{Comparison of AIO and existing methods on cross-/multi-modality retrieval.} The number in $(\cdot)$ in the ``Multi'' column indicates the number of support modalities at inference time.}
    \label{tab:comparison}
    \vspace{-6mm}
\end{table}

Meanwhile, in real-world scenarios, individuals of interest frequently encounter unknown environments that are not learned during training, \ie, zero-shot ReID in the wild. Existing methods explore domain generalizability based on a single modality, which fails to handle multi-modal zero-shot retrieval.  
Recently, foundation large models have shown their power in diverse language and vision tasks. Pioneering models such as CLIP \cite{radford2021learning} and CoCa \cite{yu2022coca}, exemplify the prowess of large-scale pre-trained foundation models as robust zero-shot performers. This characteristic holds significant relevance for ReID tasks.
Despite the existence of several large-scale ReID pre-trained models \cite{luo2021selfsupervised, yang2023unified, fu2021unsupervised}, their zero-shot performance falls short of expectations. Typical down-stream fine-tuning or training strategies would be too resource-demanding in a new challenging scenarios, \eg, data collection and annotation. Moreover, the cost of training a large-scale foundation model is too high to afford for most researchers and small companies.  
Thus, \textit{2) Is there a straightforward method to utilize extensive pre-trained foundational models for improving zero-shot performance in ReID with uncertain modalities?} 

To address the aforementioned issues, we introduce an innovative All-in-One (AIO) framework to tackle the challenges inherent in zero-shot multimodal ReID. As shown in \cref{fig:comparison}(b), the key idea of our work is to explore the potential of leveraging transformer-based foundation models to address uncertain multimodal retrieval, enhancing zero-shot ability in multimodal ReID, \eg any combination of RGB, IR, sketch, text, or simple cross-modal retrieval. 
AIO represents an experimental effort, being the \textbf{first} framework capable of simultaneously accommodating all four commonly used modalities in different ReID tasks concurrently.

In order to achieve the above goal, AIO firstly designs a lightweight multimodal tokenizer to unify diverse data. It is followed by a frozen foundation model that serves as a shared feature encoder, extracting a generalized semantic representation across all modalities and improving zero-shot performance. Then, to guide cross-modal and multimodal feature learning, AIO proposes several cross-modal heads which contain: \textit{a) A Conventional Classification Head} is utilized as the foundation of the learning guidance, learning identity-invariant representations; \textit{b) Vision Guided Masked Attribute Modeling} is introduced to learn fine-grained features and build a relationship between text and images; \textit{c) Multimodal Feature Binding} is utilized to close features of diverse modalities together. 

Furthermore, the acquisition of multimodal data in real-world scenarios poses considerable challenges, particularly concerning IR and Sketch images. The shortage of IR cameras and the substantial human labor involved in sketch drawing contribute to this difficulty. 
Existing multimodal learning methods \cite{poklukar2022geometric,wang2022bevt} demand paired multimodal, a pre-requisite not consistently met in realistic environments.
In addressing the absence of certain modalities in multimodal learning, except the proposed Multimodal Feature Binding, integrates synthetic augmentation strategies, CA \cite{ye2024channel} and Lineart \cite{von2022diffusers}, to generate synthetic IR and Sketch images respectively. CA and Lineart have been shown to effectively diminish the domain gap between RGB-IR and RGB-Sketch modalities. Their utility extends to acting as a bridge that connects feature representations of the same target across diverse modalities, thereby facilitating the reduction of the modality gap \cite{ye2024channel,voynov2023sketchguided}.

Comprehensive experiments across various zero-shot cross-modal and multimodal ReID scenarios, involving all four modalities, are conducted to evaluate the performance of the proposed framework. We also explore different foundation models and multimodal input data combinations to assess the versatility of AIO. The proposed framework demonstrates remarkable performance on multimodal ReID tasks and competitive performance on cross-modal ReID tasks without additional fine-tuning, highlighting its potential as a robust zero-shot solution for complex multimodal ReID tasks.
In summary, our contributions are three-fold:
\begin{itemize}
    \item We identify a critical limitation in existing cross-modal ReID that they lack generalizability to novel modalities, coupled with poor zero-shot performance.
    \item We design a novel multimodal ReID framework, which innovatively integrates a pre-trained foundation model and a multimodal tokenizer into ReID tasks, complemented a missing modality synthesis strategy, and three cross-modal heads to learn a unified multimodal model.
    \item We perform extensive and comprehensive experiments to demonstrate that our AIO framework is capable of handling uncertain input modalities for ReID tasks, achieving competitive performance on zero-shot cross-modal and multi-modal ReID tasks.
\end{itemize}

\section{Related Work}
\label{sec:related}
\subsection{Cross-modal ReID}
Person Re-identification (ReID) can be classified into single-modal \cite{li2022pyramidal,he2021transreid,luo2019bag,zhang2021explainable} ReID and cross-modal ReID \cite{chen2023modalityagnostic,yang2022augmented}. Specifically, cross-modal ReID considers special cases in which RGB images of the target are not available but non-RGB modalities, such as infrared (IR) \cite{ye2022dynamic,yang2022learning,ye2020dynamic,yang2023translation}, sketch \cite{gui2020learning,chen2022sketch}, or description of the person \cite{zhu2021dssl,shao2022learning,gao2021contextual,yang2023unified,li2023clipreid} could be leveraged to enlarge the application area of ReID technology. 

In real-world scenarios, the RGB image of the target may not be directly available. On the one hand, the crime often happens at night, when RGB cameras cannot capture high-quality images for ReID but IR cameras could give relatively good images of the person \cite{ye2021deep}. Thus, adopting IR images for target retrieval enters the picture. Zhang \etal \cite{zhang2022fmcnet} proposes a feature-level modality compensation network to compensate for the missing modality-specific information at the feature level to help the model learn discriminative features. Wu \etal \cite{wu2023unsupervised} leverage unlabeled data and propose a progressive graph matching method to learn a relationship between IR and RGB images to alleviate the high cost of annotation.
On the other hand, the natural language descriptions of the witness or the sketch drawn based on the textual descriptions are easy to access. Pang \etal \cite{pang2018crossdomain} first design an adversarial learning method to learn domain-invariant features cross sketch and RGB. Zhai \etal \cite{zhai2022trireid} introduce a multimodal method that combines both sketch and text as queries for retrieval. 

However, in the real world, the modality of the given target information is uncertain. 
The aforementioned works could only handle exactly two modalities, which hinders the applicability of these methods. To alleviate this problem, Chen \etal \cite{chen2023modalityagnostic} proposes a modality-agnostic retrieval method that leverages RGB, sketch, and text modalities to learn modality-specific features and fuse them based on different uni/multimodal tasks. The proposed method could handle any combination of three learned modalities, expanding usage scenarios and reducing limitations. Nevertheless, it does not take IR into account and requires a complex design that lacks scalability. 
Different from that, the architecture of the proposed AIO is simple and it is easy to be extended to more modalities.

\subsection{Multimodal Learning}
Multimodal learning methods aim to utilize the complementary properties of various modalities to learn the semantics of a task \cite{liang2023foundations}. Recently, multimodal transformers \cite{li2021align,kim2021vilt,gabeur2020multimodal} have emerged as unified models that fuse different modality inputs with token concatenation rather than extracting modality-specific and cross-modality representations. However, most multimodal learning methods \cite{botach2022endtoend,wang2022bevt,poklukar2022geometric} are designed based on the assumption of the completeness of modality for training or inference, which is not always held in the real world. To face this challenge, some researchers \cite{cai2018deep,pan2021diseaseimagespecific,wang2023multimodal,lee2023multimodal} explore building multimodal methods that could handle missing modalities. ImageBind \cite{girdhar2023imagebind} projects all features of different modalities into the same feature space and leverages contrastive learning to align all modalities to a based modality. SMIL \cite{ma2021smil} proposes to estimate the latent features of the missing modality via Bayesian Meta-Learning. GCNet \cite{lian2023gcnet} introduces graph neural network-based modules to capture temporal and speaker dependencies and jointly optimize classification and reconstruction tasks.  
Similar to previous work, the proposed AIO aspires to project features from diverse modalities into a unified feature embedding. Notably, we capitalize on the inherent capabilities of the transformer, adept at handling variable input lengths. This strategy enables the model to seamlessly accommodate inputs originating from any combination of modalities, enhancing the flexibility and adaptability of the AIO framework.

\begin{figure*}[t]
    \centering
    \includegraphics[width=\linewidth]{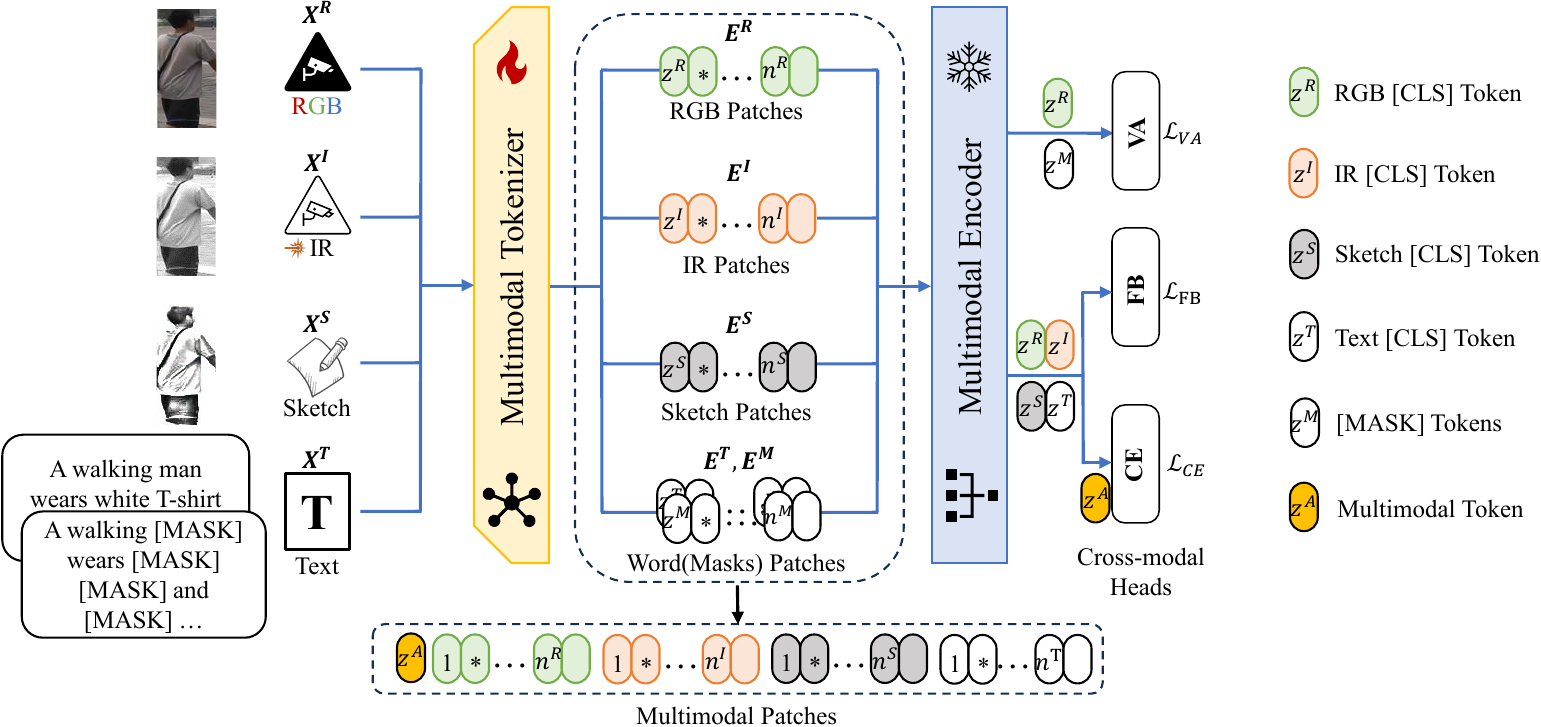}
    \vspace{-7mm} 
    \caption{\textbf{The schematic of the proposed AIO framework.} VA: Vision Guided Masked Attribute Modeling head, FB: Feature Binding head, CE: Classification head. Our framework mainly contains three parts: I) a learnable multimodal tokenizer to project diverse modalities into a unified embedding, II) a frozen foundation modal to extract complementary cross-modal representations, and III) several cross-modal heads used to dig cross-modality relationships. In order to alleviate the missing modality problem, we also leverage Channel Augmentation \cite{ye2024channel} and Lineart \cite{von2022diffusers} to synthesize IR and sketch images that are missing.}
    \label{fig:structure}
    \vspace{-5mm}
\end{figure*}

\subsection{Foundation Model}
Foundation models are designed to be adapted to various downstream tasks by pre-training on broad data at scale \cite{bommasani2022opportunities}. The efficacy of large-scale pre-trained models is evident in their capacity to enhance data encoding and elevate the performance of downstream tasks \cite{yuan2023power,huang2023federated}. Recent investigations \cite{roy2023sam, yao2022filip, ramesh2021zeroshot, marcus2022very} reveal the notable emergent ability exhibited by most large-scale pre-trained foundation models, particularly in the context of robust zero-shot performance. CLIP \cite{radford2021learning} focuses on multimodal contrastive learning on noisy web image-text pairs to learn aligned image and text representation. Impressively, CLIP achieves accuracy comparable to the original ResNet-50 on ImageNet zero-shot, without exposure to any samples from ImageNet. DALL$\cdot$E \cite{ramesh2021zeroshot} introduces a simple approach that autoregressively models sufficient and large-scale text and image tokens and demonstrates commendable zero-shot performance when compared to preceding domain-specific models. 
Nonetheless, the foundation model within ReID remains in a nascent state. The suboptimal zero-shot performance of extant large-scale pre-trained models persists due to challenges in data acquisition. Motivated by the impressive zero-shot capabilities exhibited by foundation models, the proposed AIO framework strategically employs frozen pre-trained large-scale foundation models as feature extractors, which aims to imbue our framework with the ability to learn generalized semantics from a diverse range of modalities, thereby enhancing its zero-shot performance.


\section{All in One Framework}
\label{sec:method}
In this section, we describe the proposed AIO framework in detail. AIO exhibits the capability to adeptly handle uncertain multimodal data, encompassing RGB, IR, Sketch, and Text modalities. To realize this, we introduce a novel multimodal tokenizer designed to project data into a unified embedding space. Simultaneously, a large-scale pre-trained foundation model serves as the shared feature extractor, encoding embeddings from diverse modalities. The learning process of the multimodal tokenizer is guided by cross-modal heads that are designed specifically for ReID tasks. The schematic of the AIO is illustrated in \cref{fig:structure}.

\textbf{Preliminary.}
Formally, within the AIO framework, three key components are denoted as follows: \textit{I)} the multimodal tokenizer $\boldsymbol{\psi^{mod} (\cdot)}$, \textit{II)} the frozen multimodal encoder $\boldsymbol{f (\cdot)}$, and \textit{III)} cross-modal heads $\boldsymbol{\Upsilon_{head} (\cdot)}$, where $\boldsymbol{mod}$ refers to the notation of each modality, such as RGB (R), IR (I), Sketch (S), and Text (T); $\boldsymbol{head}$ refers to the notation of Classification (CE), Vision Guided Masked Attribute Modeling (VA), and Feature Binding head (FB), respectively. 
The inputs are denoted as $\boldsymbol{x^{mod} \in X^{mod}}$, the embeddings generated by tokenizers are $\boldsymbol{E^{mod}}$, and the output feature from the frozen multimodal encoder is $\boldsymbol{z^{mod}}$, which is corresponding to class tokens in \cref{fig:structure}. 
We assume \textit{1)} each modality possesses a specific parameter space $\boldsymbol{\theta^{mod}}$ for modality-specific feature representations, and \textit{2)} there exists a shared parameter space $\theta^A$, an intersection of each modality parameter space for modality-shared feature representations, adhering to the condition:
\begin{equation}
        \theta^A \; \in \; \theta^R \cap \theta^I \cap \theta^S \cap \theta^T \text{ and } \theta^A \neq \emptyset;
\end{equation}
For a simple multimodal network with a cross-modal head $\Upsilon$ can be written as:
\begin{equation}
        \theta^A = \mathop{\arg \min}_{x \in X^{mod}}\{\Upsilon_{head} \circ z^{mod}\}
\end{equation}
where $z = f \circ \psi^{mod}(x)$, $x \in X^{mod}$ is the input from any modality, $\circ$ is the function composition operation. 
The notation of different modalities used in AIO is presented in \cref{tab:notation}.
\begin{table}\small
    \centering
    \resizebox{\linewidth}{!}{
        \begin{tabular}{c|c|c|c|c}
            \thickhline
            \rowcolor{mygray} Source & Modality & Input & Embedding & Feature\\
            \hline\hline
            \multirow{5}{*}{Raw} & RGB (R) & $X^R$ & $E^R$ & $Z^R$\\
             & IR (I) & $X^I$ & $E^I$ & $Z^I$\\
             & Sketch (S) & $X^S$ & $E^S$ & $Z^S$\\
             & Text (T) & $X^T$ & $E^T$ & $Z^T$\\
             & Multimodal (A) & - & $E^A$ & $Z^A$\\
            \hline\hline
            \multirow{3}{*}{Synthesis} & CA \cite{ye2024channel} & $X^I$ & $E^I$ & $Z^I$\\
             & Lineart \cite{von2022diffusers} & $X^S$ & $E^S$ & $Z^S$\\
             & Masked Text & $X^M$ & $E^M$ & $Z^M$\\
            \thickhline
        \end{tabular}
    }
    \vspace{-3mm}
    \caption{\textbf{Notation of different modalities.} Be aware that both Lineart \cite{von2022diffusers} and CA \cite{ye2024channel} operations are introduced, and the generated images are considered as Sketch and IR images when these two modalities are missing.}
    \label{tab:notation}
    \vspace{-5mm}
\end{table}

\subsection{Multimodal Tokenizer}
To project various modalities into a unified space, we devise a straightforward multimodal tokenizer. This tokenizer comprises four projectors: three dedicated to RGB, IR, and Sketch modalities, and one for Text. Furthermore, a multimodal embedding is constructed by amalgamating the embeddings from the respective modalities.

\noindent\textbf{Image Tokenizers.}
Given the disparate channel counts in RGB, IR, and Sketch images, for the sake of convenience, we employ channel replication in both IR and Sketch modalities to align their channel count with the three channels present in RGB images. Deviating from the original tokenizer employed in ViT \cite{dosovitskiy2021image}, leading to induce training instability \cite{wu2021cvt}, we opt for the IBN \cite{pan2018two} style tokenizer from ViT-ICS \cite{luo2021selfsupervised}. The convolutional, batch normalization (BN), and rectified linear unit (ReLU) layers inherent to the IBN-style tokenizer substantially enhance training stability \cite{caron2021emerging} and mitigate data bias which is critical for ReID \cite{luo2021selfsupervised}.

\begin{figure}
    \centering
    \includegraphics[width=\linewidth]{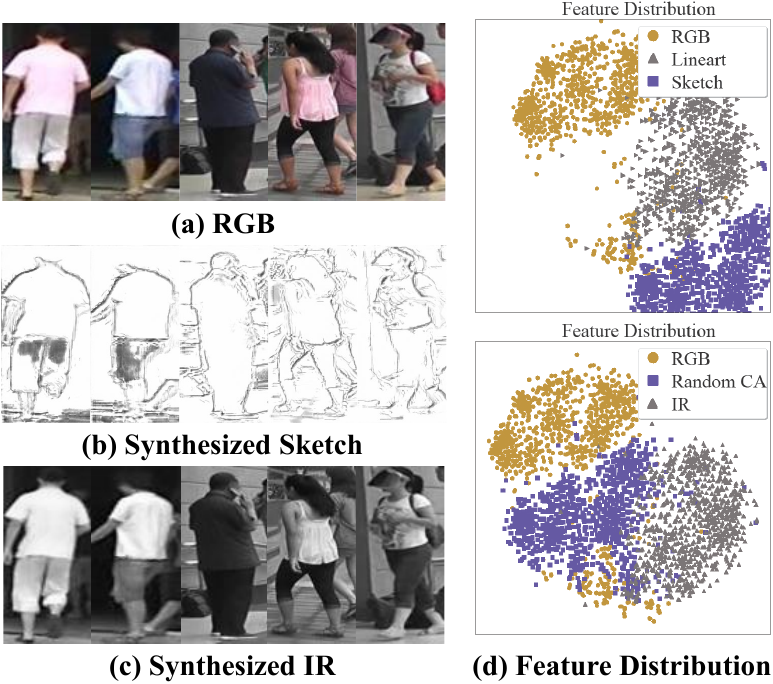}
    \vspace{-3mm}
    \caption{\textbf{The generated synthetic Sketch and IR images.} We also visualize the feature distribution of RGB, IR, Sketch, and synthesized images. }
    \label{fig:visual}
    \vspace{-5mm}
\end{figure}

\noindent\textbf{Text Tokenizers.}
In accordance with prior research efforts \cite{jiang2023crossmodal}, we adopt the CLIP tokenizer \cite{radford2021learning} to directly map the text. Each word is uniquely associated with a token, and through the utilization of word embedding layers, it is projected into a high-dimensional feature space to yield a sequence of word embeddings.

\noindent\textbf{Multimodal Embedding.}
In the context of multimodal embedding, the embeddings originating from various modalities are concatenated. Additionally, following previous works \cite{devlin2019bert,dosovitskiy2021image}, a learnable token $\boldsymbol{z^A}$ is appended to the sequence of multimodal embeddings. Simultaneously, position embeddings $\boldsymbol{E^{Pos}}$ are employed to enhance position information, seamlessly integrated with the multimodal embeddings via element-wise addition, a procedure akin to the original operation in ViT \cite{dosovitskiy2021image}. 
The multimodal embedding is formulated as follows:
\begin{equation}
    \begin{aligned}
        E^A = [z^A, E^R, E^I, E^S, E^T] + E^{Pos}, \\
        E \in \mathbf{R}^{n \times D}, E^{Pos} \in \mathbf{R}^{(n+1) \times D}.
    \end{aligned}
\end{equation}

\subsection{Missing Modality Synthesis}
\label{sec:missing}
Given the insufficiency of multimodal data in ReID, especially in IR, and Sketch, we introduce Channel Augmentation (CA) \cite{ye2024channel} and Lineart \cite{von2022diffusers} as augmentation methods to synthesize absent modalities. The generated samples are shown in \cref{fig:visual}(a)-(c).
The incorporation of synthetic modalities offers two advantages: 
\textit{1)} an expansion in the size of the multimodal sample, thereby mitigating issues associated with missing modalities; 
\textit{2)} CA and Lineart act as conduits bridging the gap between synthetic and real IR and sketch modalities. 
This is attributed to the feature distribution of the augmented images aligning between RGB and real IR and Sketch images. The visual representation of the feature distribution for RGB-Lineart-Sketch and RGB-CA-IR, as presented in \cref{fig:visual}(d) through t-SNE, serves as evidence of their efficacy in alleviating the learning challenges arising from modality gaps.

\noindent\textbf{Progressively Learning with Synthetic data.} We employ a progressively learning strategy to train the proposed AIO framework. The strategy involves initially training on synthetic images, incorporating real-world RGB and Text, for a few number of epochs. Subsequently, the model undergoes further fine-tuning using paired IR and Sketch images from the real world. This sequencing is deliberate, as synthetic images exhibit a reduced domain gap with RGB compared to real IR and Sketch images, facilitating a more accessible learning process for the model. A similar phenomenon is also found in other cross-modal works \cite{lu2023learning,zhang2023adding,ye2021visibleinfrared,zheng2024multimodal,huang2023rethinking}.

\begin{table*}\small
    \centering
    \begin{tabular}{p{1.1cm}<{\centering}|p{3.1cm}<{\raggedleft}|p{1.6cm}<{\centering}|p{1.1cm}<{\centering}|p{1.7cm}<{\centering}|p{1.3cm}<{\centering}|p{1.9cm}<{\centering}|p{1.6cm}<{\centering}}
        \thickhline
        \rowcolor{mygray} Partition & Dataset \quad\quad\quad\quad\quad & Venue & \#ID & \#RGB Imgs & \#IR Imgs & \#Sketch Imgs & \#Text \\
        \hline\hline
        \multirow{3}{*}{Train} & SYNTH-PEDES \cite{zuo2023plip} & arXiv23 & 312,321 & 4,791,771 & - & - & 12,138,157 \\
         & LLCM \cite{zhang2023diverse} & CVPR23 & 1,064 & 25,626 & 21,141 & - & - \\
         & MaSk1K \cite{lin2023domain} & ACMMM23 & 996 & 32,668 & - & 4,763 & - \\
        \hline\hline
        \multirow{5}{*}{Test} & Market1501 \cite{zheng2015scalable} & ICCV15 & 1,501 & 32,668 & - & - & - \\
         & SYSU-MM01 \cite{wu2020rgbir} & IJCV20 & 491 & 30,071 & 15,792 & - & - \\
         & PKU-Sketch \cite{pang2018crossdomain}\cite{zhai2022trireid} & ACMMM18 & 200 & 400 & - & 200 & 200 \\
         & CUHK-PEDES \cite{li2017person} & CVPR17 & 13,003 & 40,206 & - & - & 80,412 \\
         & Tri-CUHK-PEDES \cite{chen2023modalityagnostic} & CVPR23 & 13,003 & 40,206 & - & 40,206 & 80,412 \\
        \thickhline
    \end{tabular}
    \vspace{-3mm}
    \caption{\textbf{The statistics of datasets used in experiments.} More details can be found in corresponding papers.}
    \label{tab:datasets}
    \vspace{-5mm}
\end{table*}

\subsection{Multimodal Modeling and Binding}
All representations extracted by the frozen multimodal encoder from each embedding are fed into cross-modal heads $\Upsilon_{head}$, which are specifically designed to learn cross-modal relationships between different modalities. As illustrated in \cref{fig:structure}, there are three heads: \textit{1) Conventional Classification Head}, learning identity invariant representations like in other ReID works \cite{liao2020interpretable,zhuang2020rethinking,li2021weperson}; \textit{2) Vision Guided Masked Attribute Modeling}, seeking to learn fine-grained RGB-Text relationships; \textit{3) Multimodal Feature Binding}, aiming to align each modality representations together.

\noindent\textbf{Conventional Classification (CE).} The classification head only contains a bottleneck \cite{luo2019bag} and a classifier, which is constrained by Cross-Entropy loss as follows:
\begin{equation}
    \mathcal{L}_{CE} = - \frac{1}{N} \sum^N y\log(\Upsilon_{CE} \circ z^{mod}),
\end{equation}
where $N$ is the number of pedestrian IDs, $\Upsilon_{CE}$ indicates the Conventional Classification head.
The conventional Triplet Loss, commonly employed in related frameworks, is omitted in our architecture, as we opt for the utilization of a multimodal feature binding loss.

\noindent\textbf{Vision Guided Masked Attribute Modeling (VA).}
Attributes play a pivotal role in highlighting essential characteristics of an individual, encompassing factors such as gender and hair color. These attributes are instrumental in cross-modal alignment and the differentiation of distinct individuals. In this context, we investigate the utility of attribute information embedded in the Text modality to serve as supervisory signals for learning discriminative person representations.
To be specific, in the case of a paired RGB image and Text, we adopt a strategy from the prior work \cite{yang2023unified}, where specific attribute keywords in the Text are selectively masked. These masked words are then projected to a special token $[MASK]$. Subsequently, the concatenated features of the paired RGB image and the masked token are fed into a decoder structured with MLPs and a classifier, represented as follows:
\begin{equation}
    \mathcal{L}_{VA} = -\frac{1}{N_AM}\sum^{N_A}\sum^M y\log(\Upsilon_{VA}(z^R \oplus z^m)), 
\end{equation}
where $\Upsilon_{VA}$ indicates the Vision Guided Masked Attribute Modeling head, $z^m \in z^M$ are the features of masked tokens, $N_A, M$ are the number of classes and the number of masked tokens, and $\oplus$ denotes the concatenation operation. 

\noindent\textbf{Multimodal Feature Binding (FB).}
To align all modalities onto a shared manifold, we attract features from all modalities towards the RGB feature. This alignment is facilitated through the incorporation of a novel supervised feature binding loss, elucidated in the subsequent section:
\begin{equation}
    \mathcal{L}_{FB} = -\sum \log \frac{\exp(\frac{1}{mod} \sum_{mod \neq R}||z_i^R, z_i^{mod}||/\tau)}{\sum_{i \neq j} \exp(||z_i^R, z_j^R||/\tau )}, 
\end{equation}
where $|| \cdot ||$ is the cosine similarity, $z_i^R$ is the representation of person $i^{th}$ RGB embedding, $z_i^{mod}$ are the representations of person $i^{th}$ other modalities embeddings, $z_j^R$ are the RGB representations belonging to other people, $\tau$ is the temperature that controls the smoothness of the softmax distribution.
Diverging from the conventional InfoNCE approach \cite{oord2018representation}, our feature binding loss involves bringing together features from all modalities corresponding to the same individual, while simultaneously creating a separation between RGB features of distinct individuals, rather than applying the same principle to all features. This difference is motivated by the prevalence of RGB as the most common modality in real-world scenarios, contributing the most abundant data and consistently present in all publicly available datasets.

\subsection{Overall Architecture}
\label{sec:overall}
As elucidated earlier, the primary objective of the AIO framework is to learn a multimodal tokenizer through a frozen multimodal encoder, under the guidance of cross-modal heads. We believe that the emergent capabilities demonstrated in large-scale foundation models can effectively augment the zero-shot ability in multimodal ReID tasks. Additionally, capitalizing on the inherent adaptability of transformer architecture to accommodate variable input lengths, AIO exhibits competence in processing diverse combinations of commonly employed modalities in ReID. To realize this objective, AIO is constrained by three cross-modal heads, that can be written as follows:
\begin{equation}
    \mathcal{L} = \mathcal{L}_{CE} + \alpha\mathcal{L}_{VA} + \mathcal{L}_{FB},
\end{equation}
where $\alpha$ is a fixed weight to control the importance of Vision Guided Masked Attribute Modeling.

\section{Experiment}
\label{sec:exp}
In this section, we conduct a comprehensive evaluation of the proposed AIO framework across both cross-modal and multimodal ReID tasks. Our analysis demonstrates the efficacy of the AIO framework, particularly in zero-shot scenarios involving uncertain input modalities within ReID tasks. Additionally, we delve into the examination of varying foundation models and input modality combinations.
\textbf{The short notation of each modality will be used in this section}, details can be found in \cref{tab:notation}.

\subsection{Experimental settings}
\textbf{Datasets.}
Three publicly available datasets SYNTH-PEDES \cite{zuo2023plip} for R-T pairs, LLCM \cite{zhang2023diverse} for R-I images, MaSk1K \cite{lin2023domain} for R-S images are leveraged for training. 
For zero-shot performance evaluation, five widely used real-world datasets are used for evaluations, Market1501 \cite{zheng2015scalable} for R$\rightarrow$R task, SYSU-MM01 \cite{wu2020rgbir} for I$\rightarrow$R task, PKU-Sketch \cite{pang2018crossdomain} for S$\rightarrow$R task, CUHK-PEDES \cite{li2017person} for T$\rightarrow$R task, and Tri-CUHK-PEDES \cite{chen2023modalityagnostic} for T+S$\rightarrow$R task. The dataset statistics are shown in \cref{tab:datasets}. More details can be found in the original papers.

\begin{table}\small
    \centering
    \begin{tabular}{p{0.9cm}<{\centering}|p{0.9cm}<{\centering}|p{0.9cm}<{\centering}|p{0.9cm}<{\centering}|p{0.9cm}<{\centering}|p{1.1cm}<{\centering}}
        \thickhline
        \rowcolor{mygray} Heads & R$\rightarrow$R & I$\rightarrow$R & S$\rightarrow$R & T$\rightarrow$R & S+T$\rightarrow$R \\
        \hline\hline
        Base & 3.2 & 1.5 & 1.6 & 21.6 & 0.5 \\
        \hline
        + VA & 6.4 & 0.9 & 1.2 & 52.4 & 0.6 \\
        + CE & 78.0 & 43.7 & 46.8 & 53.5 & 91.8 \\
        + FB & 79.6 & 57.6 & 70.2 & 53.4 & 92.1 \\
        \thickhline
    \end{tabular}
    \vspace{-3mm}
    \caption{\textbf{Effectiveness of each cross-modal heads.} The Rank-1 zero-shot performance is reported.}
    \label{tab:heads}
    \vspace{-8mm}
\end{table}

\noindent\textbf{Evaluation Protocols.}
Following existing cross-modality ReID settings \cite{li2023clipreid,liao2022graph,wu2023unsupervised,chen2022sketch}, we use the Rank-$k$ matching accuracy, mean Average Precision (mAP) metrics, and mean Inverse Negative Precision (mINP) \cite{ye2021deep} for performance assessment.
In the context of multimodal ReID, we adhere to the evaluation settings outlined in TriReID \cite{zhai2022trireid} and UNIReID \cite{chen2023modalityagnostic} specifically designed for RGB-Text+Sketch scenarios. 
To accommodate other multimodal data combinations, we leverage CA \cite{ye2024channel} and Lineart \cite{von2022diffusers} to generate simulated IR and Sketch images. While acknowledging that this may not perfectly simulate real-world scenarios, it provides valuable insights into the multimodal performance of the proposed AIO framework.

\noindent\textbf{Implementation Details.}
We employ the ViT \cite{dosovitskiy2021image} as the backbone, which is pre-trained on LAION-2B dataset with contrastive learning, to reinforce the ability for generic token encoding. All parameters of the backbone networks are frozen. The Text tokenizer is from the pre-trained CLIP \cite{radford2021learning} to segment sentences into subwords and transform them into word embeddings. 
We perform a progressively learning strategy training process in AIO framework, as we discussed in \cref{sec:missing}. 
\textit{stage 1)} In the first 40 epochs, we sample 32 paired RGB and text samples from SYNTH-PEDES only combined with generated synthetic IR and Sketch images using CA \cite{ye2024channel} and Lineart \cite{von2022diffusers}. Moreover, we randomly chose two to four embeddings from different modalities to build the multimodal embedding. It is worth noting that, multimodal embedding may not contain RGB embedding. 
\textit{stage 2)} In the rest 80 epochs, we still select 32 samples for a batch but from all training datasets. For data from SYNTH-PEDES, the sampling, synthetic methods, and construction of multimodal embedding are unchanged. For data from LLCM and MaSk1K, only paired RGB-IR and RGB-Sketch images are leveraged. The multimodal embedding for samples from these two datasets only contains available modalities.
We also apply random horizontal flipping and random cropping for visual modalities. All images are resized to $384 \times 192$. 
The framework is optimized by AdamW \cite{loshchilov2019decoupled} optimizer with a base learning rate of 1e-4, a cosine weight decay of $1e-4$, and a warmup in the first 5 epochs. The learning rate of the CLIP tokenizer is multiplied by 1e-1 since they have already been pre-trained. The $\alpha$ is set to 3e-1 and the $\tau$ is set to 5e-2 as in \cite{girdhar2023imagebind}. The framework is distributively trained on 8 NVIDIA 3090 GPUs.

\begin{table}\small
    \centering
    \resizebox{\linewidth}{!}{
        \begin{tabular}{r|c|c|c|c|c}
            \thickhline
            \rowcolor{mygray} Model \quad\; & R$\rightarrow$R & I$\rightarrow$R & S$\rightarrow$R & T$\rightarrow$R & S+T$\rightarrow$R \\
            \hline\hline
            ViT* \cite{luo2021selfsupervised} & 74.2 & 49.6 & 63.4 & 48.2 & 79.2 \\
            Uni* \cite{li2022uniperceiver} & 76.3 & 51.3 & 68.1 & 52.6 & 84.3 \\
            CLIP* \cite{radford2021learning} & 77.7 & 55.5 & 69.5 & 52.8 & 87.6 \\
            \hline\hline
            LAION \cite{zhang2023metatransformer} $\dagger$ & 79.6 & 57.6 & 70.2 & 53.4 & 92.1 \\
            \thickhline
        \end{tabular}
    }
    \vspace{-3mm}
    \caption{\textbf{Zero-shot performance of AIO with different foundation models.} The Rank-1 performance is reported. * indicates the tokenizer of the original model is replaced by ours. $\dagger$ is the backbone used in AIO.}
    \label{tab:diff_models}
    \vspace{-3mm}
\end{table}

\begin{table}\small
    \centering
    \begin{tabular}{p{1.6cm}<{\raggedleft}|p{1.6cm}<{\centering}|p{1.6cm}<{\centering}|p{1.6cm}<{\centering}}
        \thickhline
        \rowcolor{mygray} Multimodal & Rank-1 & Rank-5 & Rank-10\\
        \hline\hline
        R+T & 56.5 & 76.2 & 85.1 \\
        R+I & 48.2 & 70.7 & 79.3 \\
        I+T & 53.4 & 74.3 & 81.0 \\
        I+S & 48.1 & 70.6 & 79.0 \\
        \hline
        R+I+T & 57.8 & 78.0 & 86.3 \\
        I+S+T & 55.6 & 74.1 & 82.2 \\
        \hline
        R+I+S+T & 58.6 & 77.9 & 86.6 \\
        \thickhline
    \end{tabular}
    \vspace{-3mm}
    \caption{\textbf{Zero-shot performance with multimodal input on Tri-CUHK-PEDES.} Be aware that the IR images are generated by using CA \cite{ye2024channel} rather than real-world IR images.}
    \label{tab:multi_input}
    \vspace{-5mm}
\end{table}

\begin{table*}\small
    \centering
    \resizebox{\linewidth}{!}{
        \begin{tabular}{c|r|c|cc|cc|cc|cc}
            \thickhline
            \rowcolor{mygray} &  &  & \multicolumn{2}{c|}{R$\rightarrow$R} & \multicolumn{2}{c|}{I$\rightarrow$R} & \multicolumn{2}{c|}{S$\rightarrow$R} & \multicolumn{2}{c}{T$\rightarrow$R} \\
            \rowcolor{mygray} \multirow{-2}{*}{Type} & \multirow{-2}{*}{Method\quad \quad} & \multirow{-2}{*}{Venue} & Rank-1 & mAP & Rank-1 & mAP & Rank-1 & mAP & Rank-1 & mAP \\
            \hline\hline
            \multirow{3}{*}{Pre-train} & LuPerson-NL \cite{fu2022largescale}  & CVPR22 & 24.6* & 11.6* & - & - & - & - & - & - \\
             & PLIP \cite{zuo2023plip}  & arXiv23 & 80.4 & 59.7 & - & - & - & - & 57.7 & - \\
             & APTM \cite{yang2023unified} & ACMMM23 & 5.3* & 3.5* & - & - & - & - & 9.6* & 2.7* \\
            \hline
            \multirow{3}{*}{Unimodal} & OSNet-IBN \cite{zhou2019omniscale} & ICCV19 & 73.0 & 44.9 & - & - & - & - & - & - \\
             & M$^3$L \cite{zhao2021learning} & CVPR21 & 78.3 & 52.5 & - & - & - & - & - & - \\
             & OSNet-AIN \cite{zhou2021learning} & TPAMI21 & 73.3 & 45.8 & - & - & - & - & - & - \\
            \hline
            \multirow{3}{*}{Cross-modal} & AGW \cite{ye2021deep} & TPAMI21 & 17.3* & 6.9* & 18.2* & 19.1* & - & - & - & - \\
             & IRRA \cite{jiang2023crossmodal} & CVPR23 & 66.6* & 40.5* & - & - & - & - & 30.1* & 25.3* \\
             & UNIReID \cite{chen2023modalityagnostic} & CVPR23 & 19.0* & 8.2* & - & - & 69.8 & 73.0 & 11.6* & 9.7* \\ 
            \hline
            Multimodal & AIO (Ours) & - & 79.6 & 59.9 & 57.6 & 51.9 & 70.2 & 73.5 & 53.4 & 43.4 \\
            \thickhline
        \end{tabular}
    }
    \vspace{-3mm}
    \caption{\textbf{Zero-shot performance on cross-modal retrieval.} The best Rank-1 and mAP performance are reported. Results with * indicate that the experiment results are produced by authors. For AGW \cite{ye2021deep}, it is trained on MSMT17 \cite{wei2018person} and LLCM \cite{zhang2023diverse} for R$\rightarrow$R and I$\rightarrow$R. For IRRA \cite{jiang2023crossmodal}, it is trained on ICFG-PEDES \cite{ding2021semantically} for T$\rightarrow$R. For UNIReID \cite{chen2023modalityagnostic}, it is trained on Tri-ICFG-PEDES \cite{chen2023modalityagnostic} for T$\rightarrow$R and R$\rightarrow$R.}
    \label{tab:cross-modal}
    \vspace{-5mm}
\end{table*}

\begin{table}\small
    \centering
    \begin{tabular}{p{2.8cm}<{\raggedleft}|p{1.2cm}<{\centering}|p{1.2cm}<{\centering}|p{1.2cm}<{\centering}}
        \thickhline
        \rowcolor{mygray} Method \quad\quad\;\; & Rank-1 & mAP & mINP\\
        \hline\hline
        UNIReID (T$\rightarrow$R) & 76.8 & 80.6 & 77.8 \\
        UNIReID (S$\rightarrow$R) & 69.8 & 73.0 & 68.3 \\
        AIO (T$\rightarrow$R) & 78.2 & 81.7 & 78.4 \\
        AIO (S$\rightarrow$R) & 69.8 & 72.8 & 68.8 \\
        \hline
        UNIReID (S+T$\rightarrow$R) & 91.4 & 91.8 & 89.0 \\
        AIO (S+T$\rightarrow$R) & 92.1 & 92.2 & 89.2 \\
        \hline
        AIO (R+S+T$\rightarrow$R) & 93.6 & 93.7 & 90.0 \\
        \hline
        AIO (R+I+S+T$\rightarrow$R) & 93.8 & 93.7 & 90.3 \\
        \thickhline
    \end{tabular}
    \vspace{-3mm}
    \caption{\textbf{Zero-shot performance with multimodal input and generalized cross-modal on PKU-Sketch.}}
    \label{tab:multimodal}
    \vspace{-5mm}
\end{table}

\subsection{Ablation Study}
\textbf{Efficacy of Designed Modules.}
We first evaluate the effectiveness of the designed components. As evident from \cref{tab:heads}, each introduced cross-modal head proves crucial for the overall performance of our All-in-One (AIO) framework. Specifically, VA head yields the most substantial performance enhancement in the T$\rightarrow$R task, CE head plays an important role in the R$\rightarrow$R task, and FB head improves all cross-modal and multimodal tasks.

\noindent\textbf{Different Foundation Models.}
We explore various foundation models, including Uni-Perceiver v2 \cite{li2022uniperceiver} (Uni), a Vision Transformer (ViT) pre-trained on LuPerson \cite{luo2021selfsupervised}, and the pre-trained image encoder from CLIP \cite{radford2021learning}. The performance of these diverse foundation models is presented in \cref{tab:diff_models}. As discernible from the table, the performance demonstrates an upward trend with the expansion of the pre-training dataset. Notably, despite LuPerson's exclusive focus on ReID tasks, its performance lags behind other models due to its comparatively smaller size. This discrepancy underscores the pronounced zero-shot performance benefits associated with large-scale pre-trained foundation models.

\noindent\textbf{Influence of Multimodality Input.}
Because our proposed AIO framework supports any combination of diverse modalities as input, we also analyze the influence of different combinations of multimodal inputs on Tri-CUHK-PEDES \cite{chen2023modalityagnostic}, where the missing IR modality is generated by CA \cite{ye2024channel}. As presented in \cref{tab:multi_input}, our analysis indicates a preference for RGB and text modalities within our framework over other modalities. Furthermore, when the number of input modalities reaches or exceeds three, there is no significant alteration in performance. This outcome aligns with expectations, as RGB and Text modalities inherently provide more discriminative details than others.

\subsection{Evaluation on Multimodal ReID}
Given the rarity of generalizable works across cross-modal, multimodal, and pre-trained ReID, we conduct a comprehensive comparative analysis involving the proposed AIO framework, various large-scale pre-trained ReID models, unimodal generalized methods, cross-modal methods, and multimodal methods, all within the zero-shot setting. 
As illustrated in \cref{tab:cross-modal}, the existing large-scale pre-trained ReID models, with the exception of PLIP, exhibit unsatisfactory performance in the zero-shot setting. 
Moreover, AIO achieves competitive performance compared to unimodal generalization methods on R$\rightarrow$R retrieval task and outperforms cross-modal methods on all cross-modal retrieval tasks in the zero-shot setting. 
Notably, existing methods fall short in generalizing to unseen modalities, a limitation overcome by AIO, which adeptly handles all four modalities in cross-modal tasks.
The outcomes presented in \cref{tab:multimodal} unveil the remarkable performance of the proposed AIO framework when incorporating multimodal input. This superior performance stands in stark contrast to methods relying solely on unimodal inputs in cross-modal tasks. Additionally, the results consistently underscore the impact of different modalities, aligning with the conclusions drawn from our preceding ablation studies that AIO is more in favor of Text and RGB modalities than others.
Moreover, we also discuss the difference between AIO and UNIReID in detail and the limitation of AIO in the supplemental part.

\section{Conclusion}
\label{sec:sconclusion}
To the best of our knowledge, this is the \textbf{first} work delving into the uncertain multimodal ReID tasks encompassing all four prevalent modalities, \eg RGB, IR, Sketch, and Text. 
We investigate the feasibility of harnessing large-scale foundation models for multimodal ReID tasks, presenting a prospective avenue toward zero-shot multimodal ReID in wild conditions. In order to cooperate with foundation models, we introduce an innovative multimodal tokenizer, designed to utilize disparate modality inputs within a shared embedding space, guided by carefully crafted cross-modal heads. Moreover, we introduce synthetic augmentation methods with a progressively learning strategy to alleviate the missing modality problem and mitigate the cross-modal gap between different modalities. Extensive experimentation demonstrates the efficacy and competitive performance of the proposed AIO framework across both zero-shot cross-modal and multimodal ReID tasks. 

\noindent\textbf{Acknowledgement.}
This work is partially supported by National Natural Science Foundation of China under Grant (62176188, 62361166629, 62225113, 62306215), and the Special Fund of Hubei Luojia Laboratory (220100015).

{
    \small
    \bibliographystyle{ieeenat_fullname}
    \bibliography{main}
}

\clearpage
\setcounter{page}{1}
\maketitlesupplementary

\section*{Differences between UNIReID and AIO}
\label{sec:diff}
There are three distinctions between UNIReID and AIO:

\noindent 1) \textbf{Divergent Goals:} UNIReID and AIO fundamentally differ in their objectives. UNIReID aims to construct a multimodal model for intra-domain retrieval with the descriptive query. At the same time, AIO is explicitly crafted for universal retrieval in real-world scenarios, with four arbitrary modalities or their combinations. Notably, all experiments in this paper follow a zero-shot generalizable setting, which is inapplicable for UNIReID.

\noindent 2) \textbf{Different Challenges:} 
UNIReID demands paired multimodal data. In comparison, AIO confronts even more challenging scenarios, involving unpaired heterogeneous multimodal data, with imbalanced and missing modalities. Thus, we introduce synthesized modalities and build connections among imbalanced modalities.

\noindent 3) \textbf{Disparate Approach:} UNIReID incorporates multiple tasks to accommodate uncertain multimodal input. The number of optimization objectives of UNIReID grows exponentially with the number of modalities, making it hard to extend to more modalities and hindering its scalability. 
Conversely, AIO designs a flexible solution, treating uncertain multimodal input as variable input lengths. It leverages the adaptable nature of the transformer architecture, simplifying the integration of additional modalities.
Furthermore, UNIReID employs separate encoders for various modalities, resulting in a lack of synergy between distinctive modalities. Different from UNIReID, AIO leverages a shared foundation model as the backbone to collaboratively learn comprehensive knowledge from heterogeneous multimodal data to complement each other and enhance its generalizablity in real-world scenarios.

\noindent All these differences make AIO more robust and generalizable than UNIReID in real scenarios.

\section*{Limitation}
\label{sec:limitation}
1) The computational complexity of AIO, necessitating $\mathcal{O}(n^2 \times D)$ operations for processing token embeddings $E^A, E^R, E^I, E^S, E^T$, particularly in the context of multimodal input, imposes a substantial memory cost and computational burden. This complexity poses challenges in scalability for incorporating additional modalities and deployment on resource-constrained edge devices. 
We assess the inference speed across varying numbers of modalities. \cref{tab:complexity} shows that the computation complexity escalates exponentially with the increase in the number of modalities, as anticipated.
\begin{table}[h]
    \centering
    \vspace{-10pt}
    \begin{tabular}{c|c}
    \thickhline
        Number of & Inference \\
        Modalities & Speed (ms) \\
        \hline\hline
        1 & 10.23 \\
        2 & 47.66\\
        4 & 181.32 \\
    \thickhline
    \end{tabular}
    \vspace{-10pt}
    \caption{\textbf{Computation complexity in the different number of input modalities.} All results are calculated with 700 samples.}
    \label{tab:complexity}
    \vspace{-12pt}
\end{table}

\noindent 2) Furthermore, it is worth noting that the implementation of multimodal ReID on synthetic data may not perfectly align with real-world scenarios, but also brings valuable insights for future works. 

\noindent 3) Moreover, the learnable parameters within the tokenizer are constrained compared to approaches that fine-tune the entire backbone, presenting a double-edged sword. While AIO is lightweight and user-friendly, it may not capture as much detailed knowledge as some alternatives. To address this challenge, a promising way is to selectively unfreeze a subset of deep layers within the backbone model, a direction we plan to investigate in future work.

\end{document}